\documentclass[10pt,twocolumn,letterpaper]{article}

\usepackage[pagenumbers]{iccv} 

\usepackage{comment}
\usepackage{amsmath}
\usepackage{amssymb}
\usepackage{amsthm}
\usepackage{algorithm}
\usepackage{algorithmic}
\usepackage{listings}
\usepackage{xcolor}
\usepackage{booktabs}
\usepackage{multirow}

\newcommand{\ours}{CART}
\newcommand{\oursindexing}{CART-Indexing}
\newcommand{\ada}{Adamerging+CART}
\newcommand{\COMMENTTRIANGLE}[1]{\textcolor{blue}{$\triangleright$ #1}}

\newcommand{\vitb}{ViT-B/32}
\newcommand{\vitl}{ViT-L/14}

\newcommand{\adata}{Adamerging+TA}

\theoremstyle{plain}

\newtheorem{theorem}{Theorem}

\definecolor{iccvblue}{rgb}{0.21,0.49,0.74}
\usepackage[pagebackref,breaklinks,colorlinks,allcolors=iccvblue]{hyperref}

\def\paperID{4496}
\def\confName{ICCV}
\def\confYear{2025}

\title{Revisiting Weight Averaging for Model Merging}

\author{Jiho Choi\textsuperscript{\rm 1\thanks{Equal contribution.}}, Donggyun Kim\textsuperscript{\rm 1\footnotemark[1]}, Chanhyuk Lee\textsuperscript{\rm 1} and Seunghoon Hong\textsuperscript{\rm 1}\thanks{Corresponding author}\\
\textsuperscript{\rm 1}KAIST, South Korea\\
{\tt\small \{jiho.choi,kdgyun425,chan3684,seunghoon.hong\}@kaist.ac.kr}}

\begin{document}
\maketitle
\begin{abstract}
Model merging aims to build a multi-task learner by combining the parameters of individually fine-tuned models without additional training. 
While a straightforward approach is to average model parameters across tasks, this often results in suboptimal performance due to interference among parameters across tasks. 
In this paper, we present intriguing results that weight averaging implicitly induces task vectors centered around the weight averaging itself and that applying a low-rank approximation to these centered task vectors significantly improves merging performance.
Our analysis shows that centering the task vectors effectively reduces task interference and most of task-specific knowledge is concentrated in the top singular vectors.
Our method demonstrates robust and scalable performance on vision benchmarks across varying numbers of tasks and model sizes. Furthermore, we observe that our approach is applicable to natural language processing tasks with competitive performance.
\end{abstract}

\section{Introduction}
\label{sec:intro}

Model merging has emerged as an efficient way to construct multi-task learners~\cite{li2023deep}.
Unlike traditional multi-task learning approaches that directly train a single model on multiple tasks~\cite{caruana1997multitask}, model merging leverages individually fine-tuned models and fuses their parameters to create a model that preserves their original capabilities.
This approach eliminates the need to prepare training data or store separate sets of parameters for each task, thereby reducing the costs associated with modern deep neural networks, which often require large amounts of data and numerous parameters.
Consequently, model merging has been favored in various applications, including federated learning~\cite{qi2024model}, model compression~\cite{huang2025emr, wang2024localizing}, and continual learning~\cite{ilharco2022patching, marczak2024magmax}.

Alongside fundamental observations about mode connectivity in parameter space~\cite{garipov2018loss,draxler2018essentially,frankle2020linear}, model merging through direct interpolation between fine-tuned parameters has become prevalent in the literature.
For example, the most straightforward approach is simply averaging the parameters of different models, a method known as weight averaging~\cite{wortsman2022model,choshen2022fusing,ilharco2022patching,matena2022merging,jin2023regmean}.
However, when the knowledge encoded in each model differs significantly, weight averaging can lead to substantial interference between parameters.
This interference often results in considerable performance degradation compared to traditional multi-task learning approaches, especially as the number of merged models increases.

To mitigate the limitations of the weight averaging, recent approaches have leveraged the task arithmetic framework~\cite{ilharco2023editing}, which allows for the extrapolation of the parameters.
In this framework, models are assumed to be fine-tuned from a common initialization, enabling the definition of \emph{task vectors} as the directions pointing from the initialization to the fine-tuned parameters in the parameter space.
Among several arithmetic operations on task vectors, addition with a scaling coefficient has shown to be effective in merging models trained on various tasks.
Subsequent approaches have improved this arithmetic by either resolving interference between task vectors~\cite{yadav2024ties, gargiulo2024task} or applying test-time adaptation techniques for better scaling coefficient~\cite{yang2024adamerging}.

In this paper, we revisit the weight averaging strategy through the lens of task arithmetic.
We begin by formulating weight averaging as a task arithmetic that induces centered task vectors around the weight average itself.
We then observe that applying low-rank approximations on the centered task vectors dramatically improves the performance of the merged model produced by the corresponding task arithmetic.
When an appropriate rank is chosen, this approach largely outperforms the original task arithmetic and even advanced task vector-based variants.
We also note that the optimal rank can be consistently found throughout different experimental settings.
To understand this surprisingly high performance, we provide both theoretical insights and empirical evidence based on the spectral analysis of task vectors.
With analysis on task interference and spectral analysis of task vectors, we show that centering the task vectors effectively reduces task interference, and most of the task-specific knowledge is concentrated in the top singular vectors. 
Based on this observation, we propose a novel training-free model merging approach called {\bf C}entered {\bf A}rithmetic with {\bf R}ank-reduced {\bf T}ask vectors ({\bf \ours{}}), which consistently outperforms existing baselines across various tasks and model architectures.
This method can be seamlessly integrated into existing merging approaches based on task vectors, such as those using test-time adaptation.
In our experiments, we evaluate \ours{} using various vision and NLP tasks and demonstrate that it significantly outperforms existing model merging approaches, both in the settings with and without test-time adaptation.
Our contributions are summarized as follows:
\begin{itemize}
    \item We propose revisiting weight averaging from the perspective of task arithmetic and present intriguing properties that emerge when combined with low-rank approximations.

    \item We conduct analysis regarding inteference and spectral aspect of task vector, revealing that centering the task vectors reduces task interference.
    \item Our extensive experiments on vision and NLP benchmarks with varying numbers of tasks demonstrate that this simple yet effective approach shows competitive performance over prior model merging methods.

\end{itemize}

\section{Related Work}
\label{sec:related_work}
\paragraph{Multi-task Learning}
Multi-task learning (MTL) aims to construct a single model that efficiently handles multiple tasks in terms of parameter and inference efficiency, typically by training the model using all available training data for each task \cite{caruana1997multitask}.
However, the increasing size of deep learning models \cite{nakkiranDeepDoubleDescent2019, villalobosMachineLearningModel2022} makes retraining computationally infeasible when new tasks are added.
Moreover, accessing the data distribution of each task is often challenging due to privacy concerns, rendering it practically difficult to utilize all training data.
While MTL opts to improve generalization and enhance the performance of each task by learning multiple tasks simultaneously, it often results in decreased performance compared to models trained on individual tasks due to task conflicts \cite{senerMultiTaskLearningMultiObjective2019, liuConflictAverseGradientDescent2024}.

\paragraph{Parameter Manipulation and Task Arithmetic}
To overcome the limitations of traditional multi-task learning, various methods have been proposed to construct a single model capable of handling multiple tasks by manipulating model parameters alone \cite{singhModelFusionOptimal2023, wortsman2022model, ilharcoPatchingOpenvocabularyModels2022}.
However, due to the non-convexity of loss functions and the complexity of the loss landscape in deep learning tasks, naively summing the parameters of models trained on different tasks often leads to significant performance degradation on each task \cite{liVisualizingLossLandscape2018, neyshaburWhatBeingTransferred2021}.

Intriguingly, \citet{ilharco2023editing} demonstrated that arithmetic operations in parameter space using \textit{task vectors}—differences between the parameters of task-specific fine-tuned models and those of a common pretrained model—consistently influence model behavior.
Specifically, adding task vectors enhances performance on multiple tasks (task addition), while subtracting a task vector diminishes performance on the associated task (task negation). This suggests that task vectors capture task-specific knowledge in a way that can be algebraically manipulated.
\citet{ortiz-jimenezTaskArithmeticTangent2023} analyzed this phenomenon, attributing it to \textit{weight disentanglement}—where different directions in the weight space of a model independently govern distinct regions in the input space, allowing the model to manipulate these regions separately for specific tasks—and described it as an emergent property inherent to pretrained parameters.

\paragraph{Model Merging and Task Interference}
Building upon the findings of \citet{ilharco2023editing}, multi-task models can be constructed without retraining or access to training data by performing a weighted sum of task vectors. However, task interference still leads to reduced performance on individual tasks.
To mitigate task interference in multi-task model merging, recent studies have proposed methods that directly operate on task vectors: Ties-Merging \cite{yadav2024ties} employs element-wise manipulations by trimming parameters with small magnitudes and resolving sign conflicts among parameters with large magnitudes via majority voting across tasks; similarly, Consensus Merging \cite{wang2024localizing} leverages task-specific binary masks to retain only those parameters deemed important by multiple tasks, thereby reducing interference-inducing elements.
Recently, \citet{yang2024adamerging} introduced a method that employs test-time adaptation \cite{wangTentFullyTesttime2021} to dynamically determine the coefficients for the weighted sum of task vectors based on the input data distribution, thereby reducing task interference.
Nonetheless, these methods are limited in further reducing task interference because they rely on the fixed definition of task vectors from the pretrained parameters.

\section{Preliminary}
\label{sec:preliminary}
\paragraph{Problem Setup}
We address the problem of model merging, which seeks to combine multiple task-specific models into a single model capable of performing all tasks. 
Let $\theta_1, \cdots, \theta_T$ denote the parameters of the models trained for $T$ different tasks and  $\mathcal{L}_t(\theta)$ represent the corresponding loss function for task $t$.
Our objective is to merge the parameters to produce $\theta_*$ that can perform all tasks, \emph{i.e.,} $\mathcal{L}_t(\theta_*) \approx \mathcal{L}_t(\theta_t)$ for each task $t = 1, \dots, T$.  
By merging models in the parameter space, we need neither an access to training data nor maintaining all individual model parameters.
We adopt the standard setting~\cite{matena2022merging,jin2023regmean,ilharco2023editing} where the models share the same architecture and are fine-tuned from a common initialization $\theta_0$, \emph{e.g.,} a pre-trained transformer~\cite{dosovitskiy2020image,radford2021learning}.

\paragraph{Weight Averaging}
A simple way to merge the fine-tuned models is to average their parameters across tasks, which is called weight averaging~\cite{wortsman2022model,choshen2022fusing,ilharco2022patching}.
\begin{equation}
    \theta_\text{avg} = \frac{1}{T} \sum_{t=1}^T \theta_t.
    \label{eqn:weight_averaging}
\end{equation}
While weight averaging has been shown to increase robustness in merging models trained for the same task with different hyper-parameters~\cite{wortsman2022model}, it often underperforms in merging models for different tasks~\cite{matena2022merging}.
Several approaches replace the simple averaging with weighted sum to consider importance factors of each task~\cite{matena2022merging,jin2023regmean}.

\paragraph{Task Arithmetic}
Task arithmetic~\cite{ilharco2023editing} proposes an alternative approach for model merging using the concept of \emph{task vectors}.
By viewing the parameters $\theta$ as a vector in the Euclidean space, a task vector $\tau_t$ is defined as a difference between the fine-tuned parameters from the pre-trained parameters: $\tau_t = \theta_t - \theta_0$.
Each task vector $\tau_t$ represents a task-specific knowledge for task $t$, thus merging the knowledge from different tasks is achieved by performing a simple arithmetic on their task vectors:
\begin{equation}
    \mathcal{A}(\lambda) = \theta_0 + \lambda \sum_{t=1}^T (\theta_t - \theta_0),
    \label{eqn:task_arithmetic}
\end{equation}
where $\lambda \in \mathbb{R}$ is a hyper-parameter controlling the contribution of the task vectors.
Note that Eq.~\eqref{eqn:task_arithmetic} reduces to the weight averaging (Eq.~\eqref{eqn:weight_averaging}) when $\lambda = \frac{1}{T}$.
With the proper choice of $\lambda$, task arithmetic deviates from the weight averaging and has been shown to improve merging performance by amplifying the task-specific knowledge encoded in the task vectors.
Several variants has been proposed to improve the task arithmetic by resolving interference between task vectors~\cite{yadav2024ties} or adaptively choosing the coefficient $\lambda$ via test-time adaptation~\cite{yang2024adamerging}.

\section{Revisiting Weight Averaging}
\label{sec:method}

In this section, we revisit the weight averaging from the perspective of task vectors.
First, we empirically show that the weight average $\theta_\text{avg}$ serves as a good initialization for defining task vectors with low-rank approximations (Section~\ref{sec:intriguing_properties}).
Then we provide an in-depth analysis of the intriguing properties (Section~\ref{sec:analysis}).
Based on these, we propose a novel model merging approach by inducing task vectors from the weight average.

\subsection{Intriguing Properties of Weight Averaging}
\label{sec:intriguing_properties}
We begin by showing that the weight averaging in Eq.~\eqref{eqn:weight_averaging} can be rewritten in the form of task arithmetic in Eq.~\eqref{eqn:task_arithmetic} as follows:
\begin{equation}
    \bar{\mathcal{A}}(\lambda) = \theta_\text{avg} + \lambda \sum_{t=1}^T (\theta_t - \theta_\text{avg}),
    \label{eqn:centered_arithmetic}
\end{equation}
Note that $\bar{\mathcal{A}}(\lambda) = \theta_\text{avg}$ always holds regardless of $\lambda$.
Eq.~\eqref{eqn:centered_arithmetic} provides a useful insight into understand weight averaging: the weight average $\theta_\text{avg}$ is \emph{itself} the result of task arithmetic by posing $\theta_\text{avg}$ as an initial point for the vectors rather than $\theta_0$.
The induced task vectors $\bar{\tau}_t = \theta_t - \theta_\text{avg}$ can be viewed as \emph{centered}, \emph{i.e.,} summing to zero, while the original task vectors from $\theta_0$ are generally uncentered.
Considering the inferior performance of weight averaging over task arithmetic, it may also seem to suggest that centering the task vectors with $\theta_\text{avg}$ is disadvantageous.

However, we observe an intriguing trend when we apply the rank reduction on the centered task vectors $\bar{\tau}_t$.
Suppose the model consists of $L$ layers and let $\theta^l \in \mathbb{R}^{m \times n}$ be the weight matrix at $l$-th layer of rank $r$.
Then we apply the centered task arithmetic defined in Eq.~\eqref{eqn:centered_arithmetic} layer-wise, with low-rank approximation on the task vectors as follows:
\begin{equation}
    \bar{\mathcal{A}}_k(\lambda) = \theta_\text{avg}^l + \lambda \sum_{t=1}^T \text{SVD}_k(\theta_t^l - \theta_\text{avg}^l), \quad \forall~ l \le L,
    \label{eqn:centered_arithmetic_lowrank}
\end{equation}
where $\text{SVD}_k(\theta^l)$ denotes low-rank approximation of $\theta^l$ with top-k singular vectors, \emph{i.e.,} $\text{SVD}_k(\theta^l) = \sum_{i=1}^k \sigma_i \mathbf{u}_i \mathbf{v}_i^\top$ where $\mathbf{u}_i$ and $\mathbf{v}_i$ denote the i-th left and right singular vectors obtained by Singular Value Decomposition (SVD), respectively.
\begin{figure}[!t]
    \centering
    \includegraphics[width=.95\linewidth]{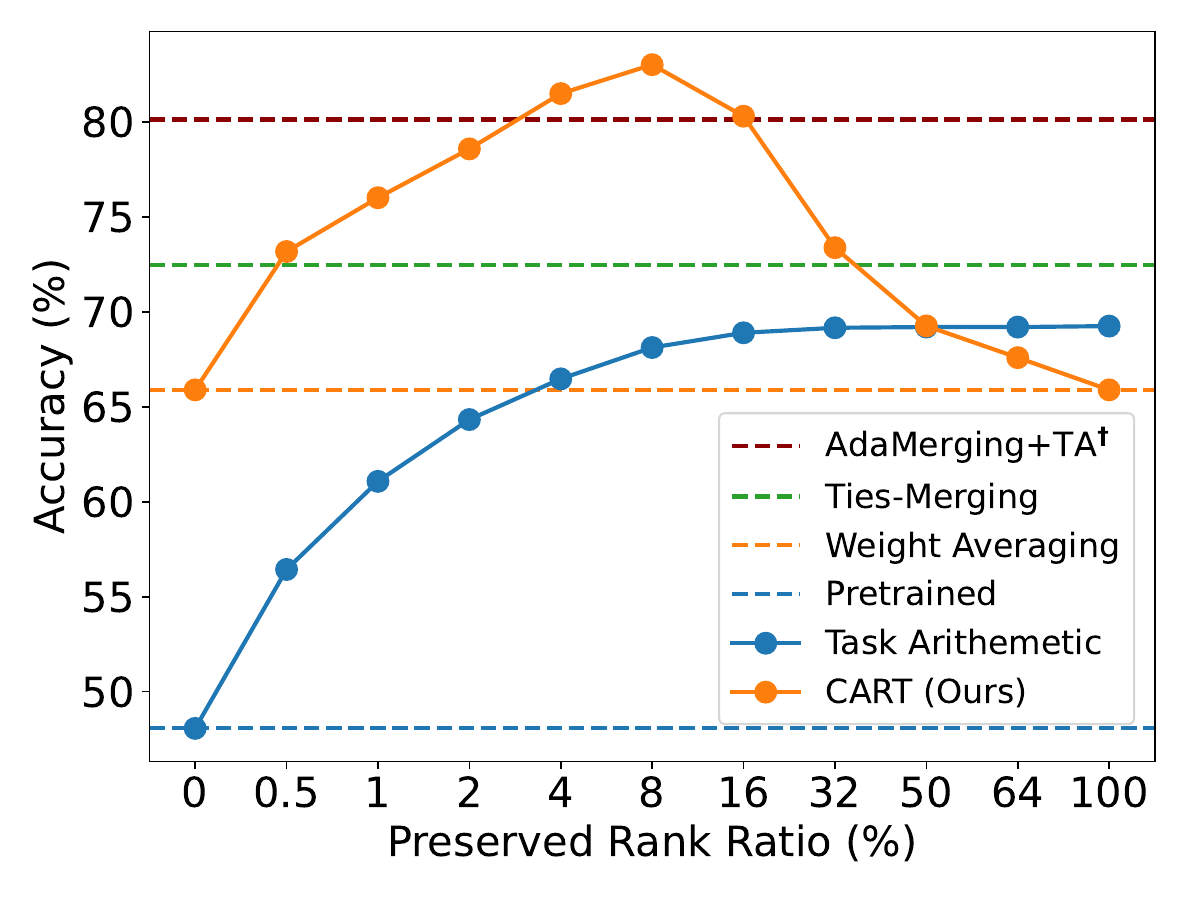}
    \caption{
        An average performance of model merging with low-rank approximations of task vectors. $\dagger$ denotes the method with test-time adaptation.
    }
    \label{fig:perf_on_ranks}
\end{figure}

Figure~\ref{fig:perf_on_ranks} illustrates the average accuracy of the merged model obtained by Eq.~\eqref{eqn:centered_arithmetic_lowrank} with varying rank $k$ (orange solid line).
As the rank increases from zero to full, we observe a rapid performance improvement beyond standard weight averaging, followed by a decline back to the weight averaging level at full rank.
The accuracies at the endpoints coincide with Eq.~\eqref{eqn:centered_arithmetic_lowrank} since it reduces to weight averaging of Eq.~\eqref{eqn:centered_arithmetic} when $k=r$ and the summation of task vectors becomes zero when $k=0$.
However, the "bell-curve" pattern at $0<k<r$ is intriguing since the peak is formed at surprisingly high accuracy, significantly outperforming the task arithmetic~\cite{ilharco2023editing}, its advanced variants such as Ties-Merging~\cite{yadav2024ties} and even the one with test-time adaptation such as AdaMerging~\cite{yang2024adamerging}.
As we show in the experiment (Figure~\ref{fig:perf_on_ranks_combined_vertical}), this trend is consistently observed across different tasks and model sizes, with the optimal performance achieved around the same rank ($\approx8\%$).

Notably, such properties are only observed when we perform the task arithmetic with centered task vectors $\bar{\tau}_t$.
When we apply rank reduction on the original task vectors $\tau_t$, the task arithmetic performance consistently decreases as rank reduces, collapsing to the pre-trained parameters at rank zero (blue solid line).
This motivates us to further investigate the task vectors through the spectral components.

\begin{figure}[!t]
    \centering
    \includegraphics[width=.95\linewidth]{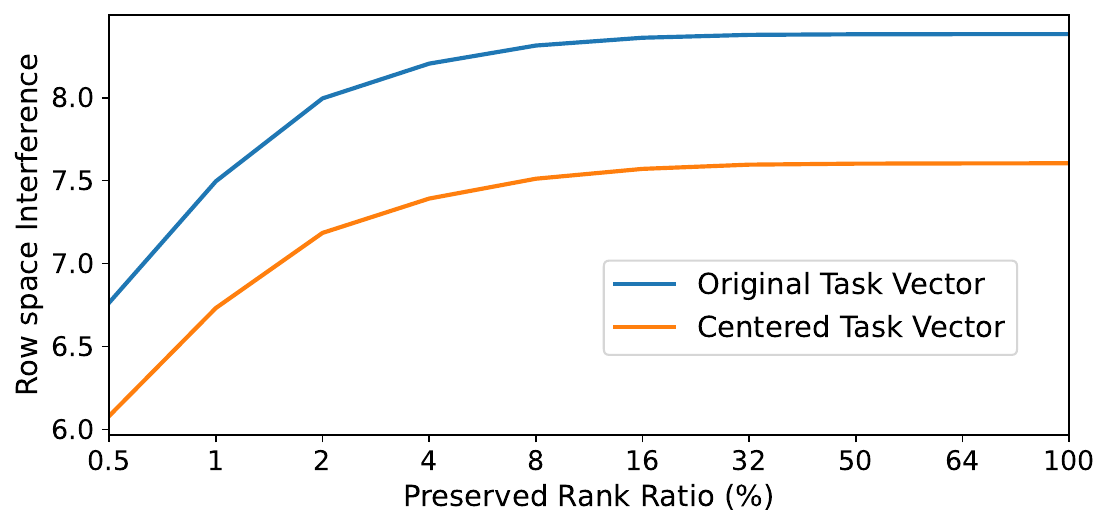}
    \caption{For the \vitb{} model with 8 vision tasks, we compute the row space interference $I(k)$ for both centered and original task vectors. The plots illustrate the results for a layer, showing that the centered task vectors consistently exhibit lower $I(k)$ values compared to the original task vectors across all ranks $k$. Layer-wise means and detailed distribution plot for each layer is provided in the Appendix~\ref{sec:anal I}.}
    \vspace{-0.5cm}
    \label{fig:task_arithmetic_reversed}
\end{figure}

\begin{figure}[!t]
    \centering
    \includegraphics[width=.95\linewidth]{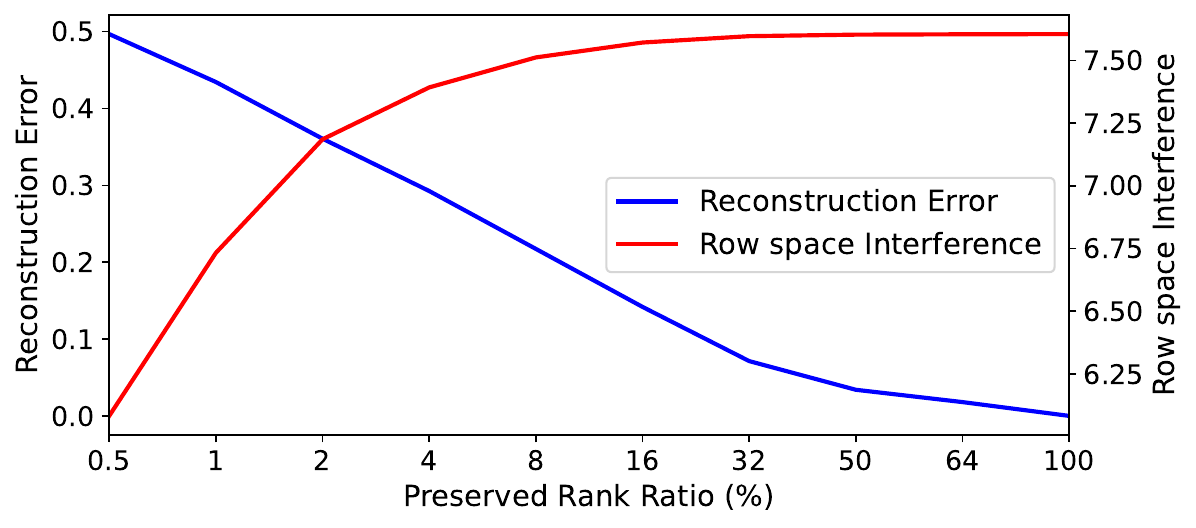}
    \caption{For the \vitb{} model with 8 vision tasks, the reconstruction error $R(k)$ and row space interference $I(k)$ are computed for the centered task vectors. In the results obtained for a layer, it is observed that as the rank $k$ increases, $R(k)$ exhibits a sharp decline in the low-rank regime. Concurrently, $I(k)$ demonstrates a gradual rise with increasing rank.}
    \vspace{-0.5cm}
    
    \label{fig:spectral_analysis}
\end{figure}

\subsection{Analysis on Singular Vectors}
\label{sec:analysis}
To understand the intriguing properties of weight averaging discussed in Section~\ref{sec:intriguing_properties}, we first examine in depth the relationship between interference among task vectors in the perspective of orthogonality.  
Based on the intuition that task interference is minimized when each task vector is spanned by nullspace of the other task vectors, we define \emph{Row space Interference} $I(k)$ as follow:
\begin{align}
    I(k) := \sum_{i=1}^{T} \sum_{\substack{j=1 \\ j \neq i}}^{T} 
    \left\| \left( \tilde{\Sigma}_{i} \tilde{V}_{i} \right)^T 
    \left( \tilde{\Sigma}_{j} \tilde{V}_{j} \right) \right\|_{F}
\end{align}
where $\tilde{V}_t$ is the top-$k$ right singular vectors of the task vector $\tau_t$, and $\tilde{\Sigma}_t$ is a diagonal matrix with corresponding singular values normalized by their $\ell_2$ norm.
Then, for a single layer, we can derive a simple theorem establishing the relationship between $I$ and task interference as follows:

\begin{theorem}
\label{thm:interference}
Assume a multi-task model defined as $\theta_{\mathrm{MTL}} = \theta_0 + \sum_{t=1}^T \tau_t$ and inputs $x_{t,i}$ to each task $t$ lie close to the subspace spanned by row space of $\tau_t$. 
If we define Task Interference $L := \sum_{t=1}^T \sum_{i=1}^{n} \| \theta_{\mathrm{MTL}} x_{t,i} - (\theta_0 +\tau_{t}) x_{t,i} \|_2^2$, then following bound holds: \begin{align}
    L \le \mathcal{O}(I^{2})
\end{align}
\end{theorem}
A more strict definition and the proof are provided in Appendix~\ref{ref: orthogonality}.
Ideally, achieving $L = 0$ implies that the input for each task is transformed exclusively by its own task vector without being influenced by other tasks. Consequently, a smaller value of $I(k)$ implies reduced interference among tasks.
In this context, Figure~\ref{fig:task_arithmetic_reversed} illustrates that, across all ranks, the centered task vectors consistently exhibit lower \( I(k) \) compared to the original task vectors. 
This observation provides insight regarding the superior performance of centered task vectors in Figure~\ref{fig:perf_on_ranks}, since reducing Row Space interference $I$ lowers Task interference by Theorem~\ref{thm:interference}.
To further understand the phenomenon that centered task vectors regress toward the average as the rank increases, we define \emph{Reconstruction Error} \( R(k) \) as: \begin{align}
R(k) := \sum_{i=1}^{T} \left\|\, \theta_{i} - \theta_{avg} - \text{SVD}_{k}(\theta_{i} - \theta_{avg}) \,\right\|_{\text{F}}^{2}.
\end{align}
Ideally, \( R(k) = 0 \) implies perfect reconstruction of the corresponding task vectors. We examine \( R(k) \) and \( I(k) \) together in Figure~\ref{fig:spectral_analysis}, observing that \( R(k) \) exhibits a steep decline in the low-rank regime. 
It indicates that a small number of top singular vectors reconstruct the centered task vectors sufficiently with low task interference. 
However, beyond this low-rank regime, the improvement in \( R(k) \) becomes marginal and the increase in \( I(k) \) begins to dominate. 
Ultimately, this interaction elucidates the regression of the centered task vectors toward the average, explaining the "bell-curve" shape of Figure~\ref{fig:perf_on_ranks}.

\subsection{Model Merging with \ours{}}
\label{sec:our_approach}

Based on the previous discussions, we propose a training-free model merging approach called {\bf C}entered {\bf A}rithmetic with {\bf R}ank-reduced {\bf T}ask vectors ({\bf \ours{}}).
The method is simple; given $T$ parameters $\theta_1, \cdots, \theta_T$ individually trained for each task $t$, we apply  Eq.~\eqref{eqn:centered_arithmetic_lowrank} to obtain merged weight matrices of the model.
Compared to the original task arithmetic, \ours{} introduces an additional hyperparameter $k$.
However, as we discuss in the experiment (Figure~\ref{fig:perf_on_ranks_combined_vertical}), retaining 8\% of the rank yields stable performance across different settings.

Thanks to the generality, \ours{} can be plugged in any merging method that leverages the task arithmetic framework.
For example, AdaMerging~\cite{yang2024adamerging} exploits test-time adaptation to improve the performance.
This introduces task-wise and layer-wise merging coefficients $\lambda_t^l$ and adapts them during test time by minimizing the Shannon entropy on the test data.
Combined with this variant, Eq.~\eqref{eqn:centered_arithmetic_lowrank} can be transformed into
\begin{equation}
    \tilde{\mathcal{A}}_k = \theta_\text{avg}^l + \sum_{t=1}^T \lambda_t^l \cdot  \text{SVD}_k(\theta_t^l - \theta_\text{avg}^l), \quad \forall~ l \le L,
    \label{eqn:centered_arithmetic_lowrank_tta}
\end{equation}
which we refer as \ada{} in our experiments.
\section{Experiments} \label{sec:experiments}
In this section, we present our experimental results.
First, we describe the experimental setup in Section~\ref{sec:exp_setup}, followed by a comparison of our method with state-of-the-art model merging techniques in Section~\ref{sec:main_results}.
Finally, we provide an in-depth analysis of our method, exploring its unique behavior, scalability, and sensitivity in Section~\ref{sec:additional_results}. 
To ensure reproducibility of our results, we release the source code.\footnote{\url{https://github.com/JH-GEECS/CART_public}}

\subsection{Experimental Setup} \label{sec:exp_setup}

\paragraph{Computer Vision}
To systematically evaluate the scalability and generalization of our model merging approach, we conducted experiments across diverse visual classification tasks with varying domains and label distributions. Initially, we considered an 8-task configuration comprising Cars~\cite{krause20133d}, DTD~\cite{cimpoi2014describing}, EuroSAT~\cite{helber2019eurosat}, GTSRB~\cite{stallkamp2011german}, MNIST~\cite{lecun1998mnist}, RESISC45~\cite{cheng2017remote}, SUN397~\cite{xiao2016sun}, and SVHN~\cite{netzer2011reading}. To further examine scalability, we expanded this setting to 14 tasks by incorporating CIFAR100~\cite{krizhevsky2009learning}, FER2013~\cite{khaireddin2021facial}, Flowers102~\cite{nilsback2008automated}, OxfordIIITPet~\cite{parkhi2012cats}, PCAM~\cite{veeling2018rotation}, and STL10~\cite{coates2011analysis}. Subsequently, we extended our evaluation to a 20-task setup by adding EMNIST~\cite{cohen2017emnist}, CIFAR10~\cite{krizhevsky2009learning}, Food101~\cite{kaur2017combining}, FashionMNIST~\cite{xiao2017/online}, RenderedSST2~\cite{socher-etal-2013-recursive}, and KMNIST~\cite{clanuwat2018deep} following prior practice~\cite{wang2024localizing}.

\paragraph{Baseline Methods}
we categorize our baselines into two groups.
The first group consists of non-merging approaches, including zero-shot application of the pre-trained model (Pretrained) and individually fine-tuned models for each task (Individual) which respectively represent the lower- and upper-performance bounds for model merging.
The second group includes Weight Average and task vector based approaches such as Task Arithmetic~\cite{ilharco2023editing}, Ties-Merging~\cite{yadav2024ties} and Consensus Merge~\cite{wang2024localizing} that aims to reduce task interference. Also, we additionally consider the method with test-time adaptation that adapts the merging coefficient layer-wise such as AdaMerging~\cite{yang2024adamerging}.

\paragraph{Natural Language Processing}
To further assess the generalizability of our method beyond the vision modality, we additionally performed experiments on eight natural language processing (NLP) classification tasks from the GLUE benchmark~\cite{wang2018glue}. Specifically, we merged fully fine-tuned task-specific models built upon the RoBERTa-base backbone~\cite{liu2019roberta}, preserving the individual task-specific classification heads. This approach enables task-wise evaluation of the merged backbone. We compared our method against task arithmetic-based model merging techniques, including weight averaging, task arithmetic~\cite{ilharco2023editing}, and Ties-Merging~\cite{yadav2024ties}. For each of these approaches, we report reproduced results using the hyperparameter settings suggested in their original papers.

\paragraph{Implementation Details}
When merging the model, we applied the low-rank approximation of Eq.~\eqref{eqn:centered_arithmetic_lowrank} to only the matrix component of the parameters such as weight matrices in MLP and project matrices in attention layers, while non-matrix components such as biases or ones in normalization layers are set to standard weight averaging. \begin{table*}[ht]
    \small
    \centering
    \caption{Multi-Task Performances on 8, 14 and 20 vision tasks with merged \vitb{} and \vitl{}.}
    \begin{tabular}{ccccccc}
      \toprule
      \multirow{2}{*}{\textbf{Method}}         & \multicolumn{3}{c}{\vitb{}}                     & \multicolumn{3}{c}{\vitl{}}                    \\ 
      \cmidrule(lr){2-4} \cmidrule(lr){5-7}
                                               & 8 tasks & 14 tasks          & 20 tasks           & 8 tasks & 14 tasks          & 20 tasks          \\ 
      \midrule
      \multicolumn{1}{c}{Pretrained}            & $48.0$ & $57.1$ & $56.0$ & $65.0$ & $68.4$ & $65.4$ \\
      \multicolumn{1}{c}{Individual}            & $90.5$ & $89.6$ & $90.5$ & $94.2$ & $93.3$ & $94.1$ \\
      \midrule
      \multicolumn{7}{c}{Without Test Time Adaptation} \\
      \midrule
      \multicolumn{1}{c}{Weight Averaging}      & $65.9$ & $64.3$ & $61.0$& $79.6$ & $76.8$ & $71.7$ \\
      \multicolumn{1}{c}{Task Arithmetic~\cite{ilharco2023editing}}       & $69.1$ & $65.4$& $60.5$ & $84.5$ & $79.6$ & $74.2$ \\
      \multicolumn{1}{c}{Ties-Merging~\cite{yadav2024ties}}          & $72.4$ & $65.2$& $62.9$ & $86.1$ & $79.5$& $75.8$\\
      \multicolumn{1}{c}{Consensus TA~\cite{wang2024localizing}}        & $75.2$ & $70.0$& $65.0$& $86.6$& $81.9$& $78.8$\\
      \multicolumn{1}{c}{\textbf{\ours{} (Ours)}}                 & $\mathbf{84.7}$ & $\mathbf{79.5}$& $\mathbf{76.3}$& $\mathbf{92.6}$ & $\mathbf{88.7}$& $\mathbf{87.9}$\\
      \midrule
      \multicolumn{7}{c}{With Test Time Adaptation} \\
      \midrule
      \multicolumn{1}{c}{\adata{}~\cite{yang2024adamerging}}            & $80.1$ & $76.7$ & $69.2$ & $90.8$& $88.0$ & $86.8$ \\
      \textbf{\ada{} (Ours)}               & $\mathbf{85.8}$ & $\mathbf{82.3}$ & $\mathbf{82.7}$ & $\mathbf{93.1}$ & $\mathbf{90.4}$ & $\mathbf{91.3}$ \\
      \bottomrule
    \end{tabular}
    \label{tab:task_specific}
  \end{table*}
\begin{figure*}[!ht]
    \centering
    \includegraphics[width=1\linewidth]{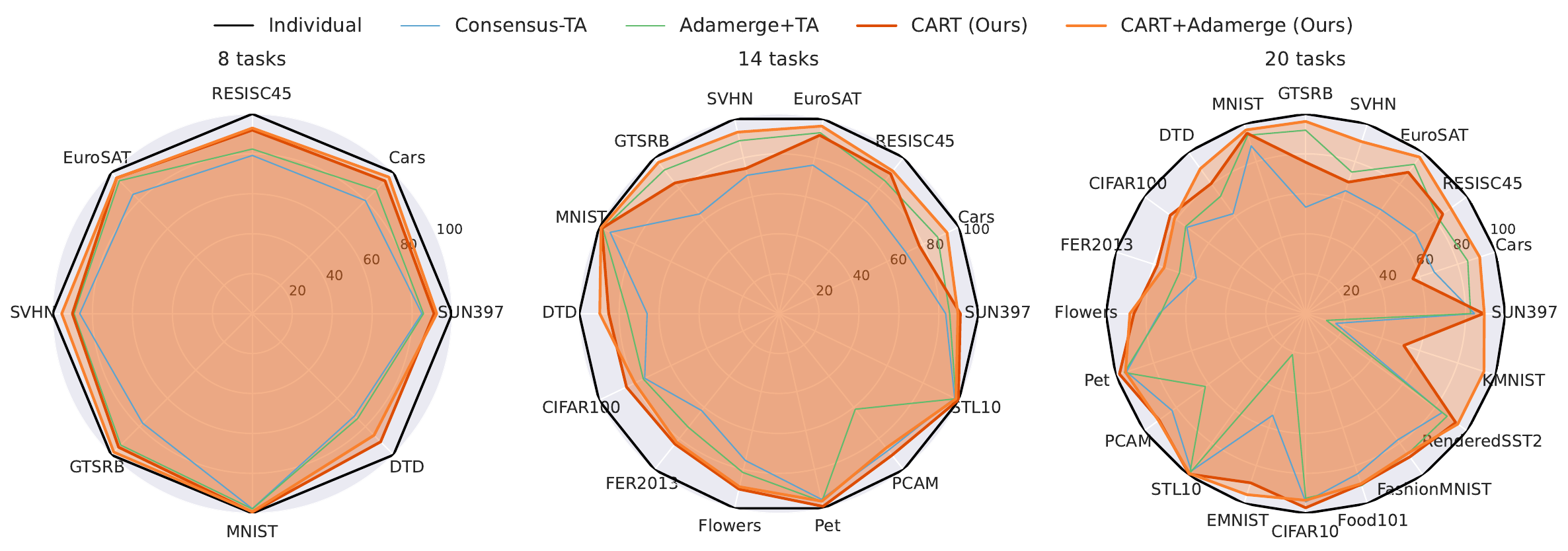}
    \caption{RADAR plot illustrating performance on sets of 8, 14, and 20 vision classification tasks for \vitb{}, with individual task accuracies normalized by their respective single-task performance. Corresponding results for \vitl{} are presented separately in Appendix~\ref{sec:large_model}.}
    \label{fig:radar_base}
\end{figure*}

\subsection{Main Results} \label{sec:main_results}
\paragraph{Computer Vision}
Table~\ref{tab:task_specific} presents the comparison with various model merging baselines.
First, it shows that the performance of naive weight averaging is far below the upper bound (Individual) and generally performs worse compared to the approaches based on task vector (Task Arithmetic).
However, we observe that applying the proper low-rank approximation to weight averaging by our method significantly boosts the performance, surpassing the prior art based on task vector (Consensus TA) up to 9.5\% for 8 tasks.
Considering that Consensus TA employed carefully hand-designed strategies to reduce conflicts among tasks, the improvement suggests that our method can be more effective in resolving inter-task conflicts.
Furthermore, this performance advantage persists consistently across experiments involving 14 and 20 tasks, thereby demonstrating the robustness of our proposed methodology with respect to varying numbers of merged tasks.

Our method outperforms strong task arithmetic approaches, including those employing test-time adaptation such as AdaMerge+TA, which optimizes layer-wise coefficients of task vectors at test time. Even without additional techniques, our method achieves superior performance across all numbers of tasks and backbone sizes. When combined with the test-time adaptation (Eq.~\eqref{eqn:centered_arithmetic_lowrank_tta}), our method exhibits further improvement.
This result indicates that our method can be effectively combined with various merging techniques within the task arithmetic framework, suggesting that it may further close the performance gap with individually finetuned models when integrated with more advanced techniques.

Table~\ref{tab:task_specific} also presents the results using a larger backbone (\vitl{}).
We observe even more pronounced gains across all model merging approaches, suggesting that model merging becomes increasingly effective with larger backbones. Consistent with previous observations, we also notice that our method significantly outperforms all baselines, both with and without test-time adaptation. Notably, applying test-time adaptation to our method (\ada{}) further narrows the gap to individually fine-tuned models. Specifically, with test-time adaptation, the performance gap for the 8-task setting shrinks to approximately 1\%, and for the 14- and 20-task settings, the gap remains within 3\%. These results demonstrate that the effectiveness of our method remains consistent across different backbone sizes and various numbers of tasks.

\begin{figure*}[ht]
    \centering
    \begin{minipage}{\linewidth}
        \small
        \centering
        \captionof{table}{Multi-Task Performance on eight NLP Tasks with Merged RoBERTa-base with CART.}
        \begin{tabular}{l | c c c c c c c c | c}
            \toprule
            Method           & CoLA   & SST2   & MRPC   & STSB   & QQP    & MNLI   & QNLI   & RTE    & \textbf{Avg.} \\
            \midrule
            Individual       & 0.6018 & 0.9404 & 0.8922 & 0.9063 & 0.9141 & 0.8720 & 0.9271 & 0.7906 & 0.8556        \\
            \midrule
            Weight Averaging & 0.1396 & 0.6411 & 0.6936 & 0.3184 & 0.7536 & 0.4219 & 0.5870 & 0.5523 & 0.5134        \\
            Task Arithmetic  & 0.1878 & 0.8589 & 0.7990 & 0.7403 & 0.8378 & 0.5908 & 0.6967 & 0.6209 & 0.6665        \\
            Ties-Merging     & 0.2048 & 0.8440 & 0.8113 & 0.5819 & \textbf{0.8570} & \textbf{0.6465} & 0.7481 & 0.4296 & 0.6404        \\
            \textbf{\ours{} (Ours)} & \textbf{0.3294} & \textbf{0.9243} & \textbf{0.8284} & \textbf{0.7956} & 0.8225 & 0.5137 & \textbf{0.7985} & \textbf{0.6498} & \textbf{0.7078}        \\
            \bottomrule
        \end{tabular}
        \label{tab:roberta}
    \end{minipage}
    \begin{minipage}{\linewidth}
        \centering
        \includegraphics[width=1.0\linewidth]{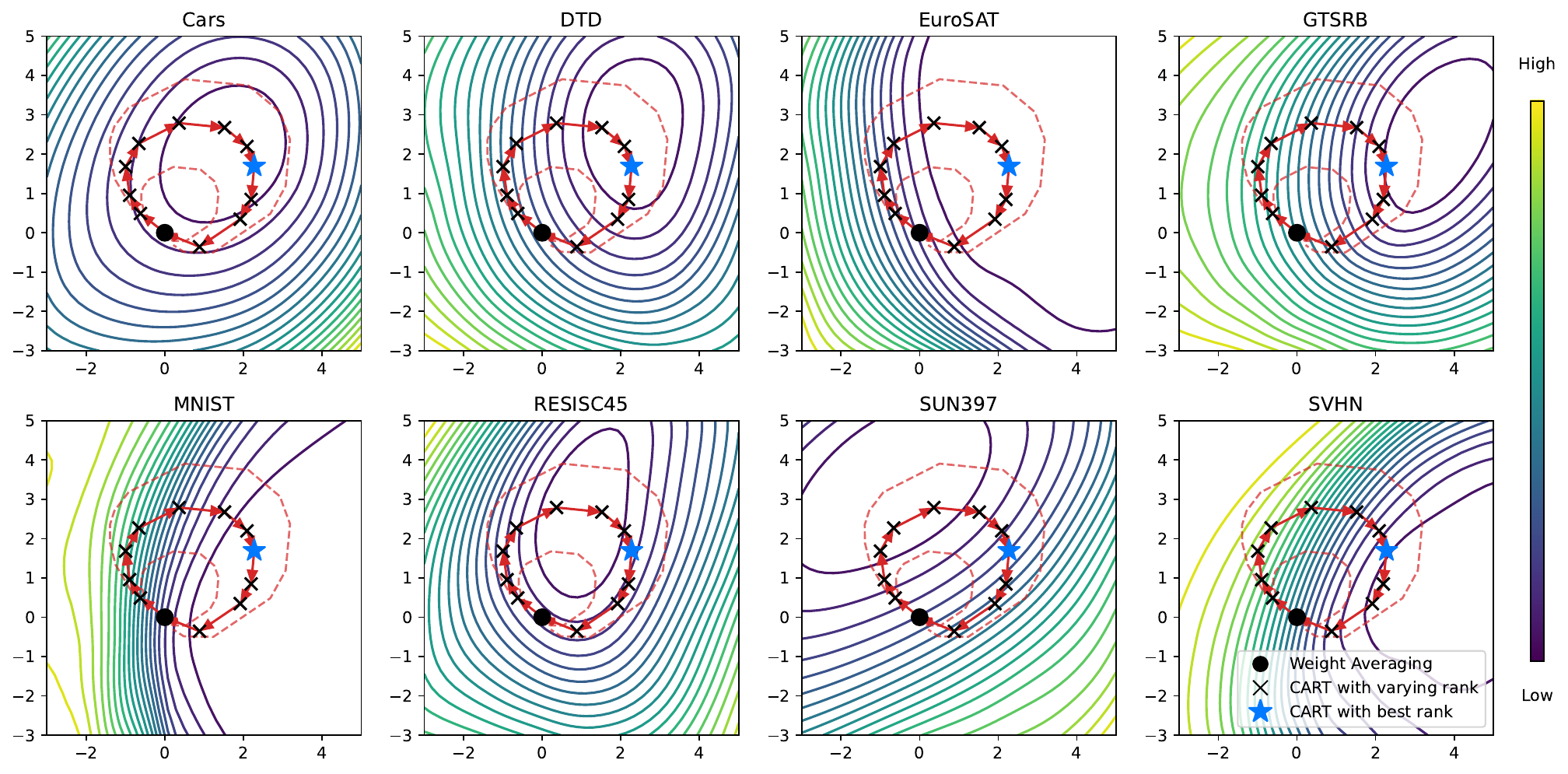}
        \caption{Loss landscape visualization of \ours{} with varying rank $k$ and the scaling coefficient $\lambda$ for \vitb{} model. We plot the trajectory of the parameters as we increase $k$ from zero to full rank, whose direction is represented as red arrows.
            The black circle indicates the weight average $\theta_\text{avg}$ and the blue star indicates \ours{} with 8\% of the full rank.
            \ours{} with other ranks are indicated as black crosses.
            Trajectories with different $\lambda$ are plotted as red dashed lines.}
        \label{fig:loss_landscapes}
    \end{minipage}
\end{figure*}

\paragraph{Natural Language Processing}
Table~\ref{tab:roberta} compares our results against various model merging baselines in the NLP domain. Similar to our findings in the vision tasks, naive weight averaging performs worse than approaches based on task vectors. However, when we apply our low-rank approximation method to weight averaging, performance improves substantially. This improvement suggests that our approach can be applied to various modalities.

\subsection{Analysis} \label{sec:additional_results}
\begin{figure*}[ht]

    \centering
    \includegraphics[width=\textwidth]{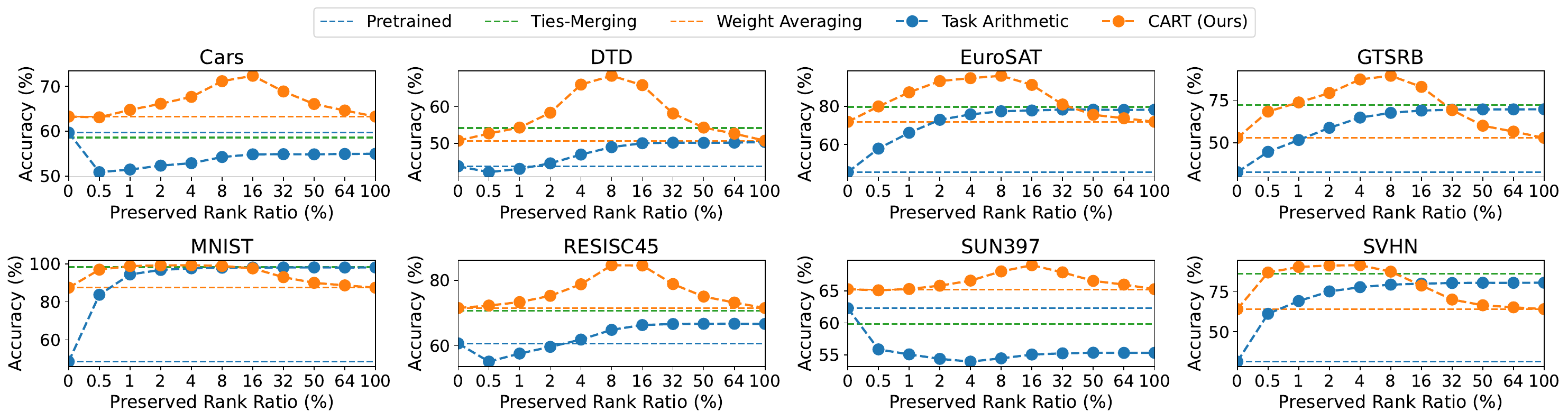}
    \caption{
        Performance evaluation of \ours{} and Task Arithmetic across eight vision tasks with low-rank approximation on their task vectors. The plots depict the accuracy (\%) of each task as a function of the preserved rank ratio (\%) of \vitb{}. (\vitl{}'s results are provided Appendix~\ref{sec:large_model})
        \ours{} consistently outperforms Task Arithmetic at a specific rank ratio (\emph{e.g.,} 8\%), demonstrating its effectiveness in the model merging process.
    }
    \vspace{-0.5cm}
    \label{fig:perf_on_ranks_combined_vertical}
\end{figure*}

\paragraph{Loss Landscape Visualization}
To investigate the effect of rank reduction in \ours{}, we visualize the loss landscapes around the weight average $\theta_\text{avg}$. 
Following \citet{li2018visualizing}, we vectorize parameters from all layers into a single flat vector, then project them onto a 2D space spanned by the first two principal components of the parameters.
Figure~\ref{fig:loss_landscapes} shows contour plots of the loss landscapes for each task, illustrating the trajectory of \ours{} as the rank $k$ from zero to full.
Consistent to Figure~\ref{fig:perf_on_ranks_combined_vertical}, the trajectory forms a circular path revolves around the weight average, visiting the loss \emph{basins} of each task.
As discussed in Section~\ref{sec:analysis}, the top and lower singular vectors of the centered task vectors play distinct roles in the parameter space.
We observe a common basin of all tasks near the weight average (located in the top-right area of each plot), where the top singular vectors of the centered task vectors guide $\theta_\text{avg}$ toward this common basin.
Conversely, the lower singular vectors introduce noise that causes the parameters to deviate from the common basin.
Combined with the scaling coefficient $\lambda$, which controls the radius of the trajectory, \ours{} enables more dynamic exploration of the loss landscapes than the linear movements of original task arithmetic, leading to a significant improvement in performance.

\begin{figure}[!t]
    \centering
    \includegraphics[width=0.9\linewidth]{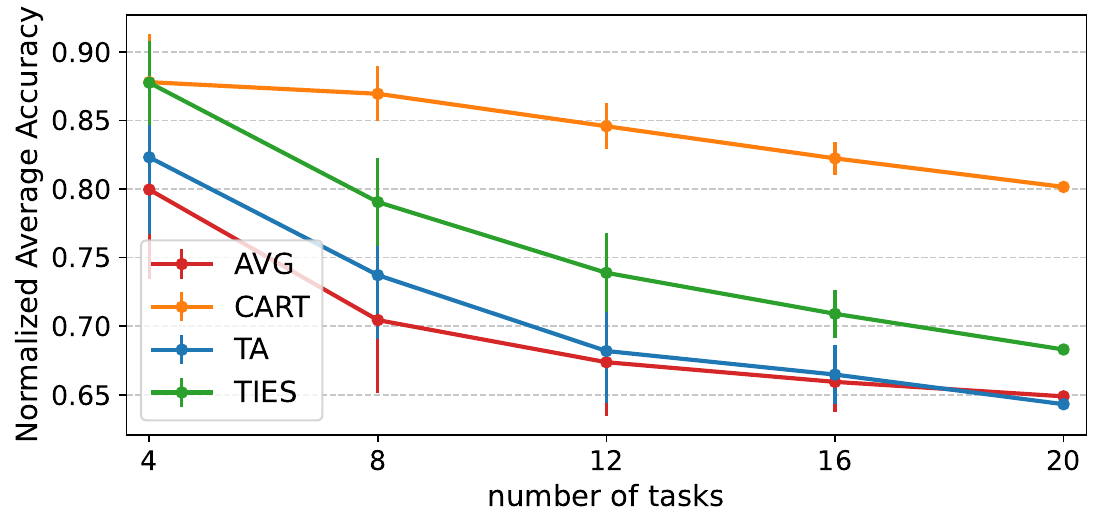}
    \caption{
        Normalized average performance versus number of tasks for different merging methods applied to \vitb{}, evaluated on sampled subsets from 20 vision tasks. Mean and standard error are reported.
    }
    \vspace{-0.65cm}
    \label{fig:performance_diff_number_tasks}
\end{figure}

\paragraph{Scalability on Number of Tasks}
In this experiment, we assess the impact of the number of tasks $N$ on model merging performance. Given the large value of $\binom{20}{N}$, we sampled multiple subsets of $N$ tasks from twenty vision tasks using a statistically sound approach in Appendix~\ref{sec:task_sampling}. This sampling, supported by the central limit theorem, ensures a reliable estimate of average performance across all possible combinations. We computed the average normalized performance over these subsets, relative to fine-tuned models, as $N$ increases. Results are presented in Figure~\ref{fig:performance_diff_number_tasks}.

We observe that methods such as simple averaging (AVG) and task arithmetic (TA) show significant performance degradation as the number of tasks increases. Similarly, Ties-Merging, despite its sophisticated parameter selection strategy, also experiences a notable decrease in performance. In contrast, our method, \ours{}, consistently achieves superior performance across all evaluated numbers of tasks. Moreover, \ours{} demonstrates stability across various task combinations, as indicated by its consistently smaller error bars compared to other methods for all task counts, with the exception of the four-task case where \ours{} and Ties-Merging exhibit comparable error bars. This robustness with respect to the number of tasks and different task combinations can be attributed to \ours{}’s effective reduction of task interference through the centering of task vectors, as detailed in Section~\ref{sec:analysis}.

\paragraph{Optimal Rank for Low-Rank Approximation}
Since the performance of our method depends on the low-rank approximation of the task vector, we investigate the impact of rank ($k$ in Eq.~\eqref{eqn:centered_arithmetic_lowrank}) in merging performance.
Figure~\ref{fig:perf_on_ranks_combined_vertical} presents the per-task performance with varying ranks on \vitb{}. 
The results with \vitl{} is presented in the Appendix~\ref{sec:large_model} (Figure~\ref{fig:perf_on_ranks_task_specific_large}).
Although the overall performance depends largely on the rank of task vectors, we observe that the optimal performance lies around $8\%$ of total dimensionality consistently over different tasks and models.
It suggests that our method is relatively robust to hyperparameter tuning, reducing the need for extensive searches to find the optimal rank.

\paragraph{Extension to model compression}
While our approach primarily focuses on model merging \emph{i.e.}, maintaining a single merged model, we can extend it to a model compression that preserves task-specific parameters. 
In this case, our method also exhibits advantages in terms of compression efficiency by representing each task vector with a small number of rank-1 matrices.
In Appendix~\ref{sec:compression}, we provide a comparative analysis for compression and show that it achieves competitive performance to strong baselines~\cite{huang2025emr,wang2024localizing}.

\section{Conclusion} \label{sec:conclusion}
In this study, we revisit the effectiveness of weight averaging from the perspective of model merging.
Through a comprehensive analysis of singular components and loss landscape, we confirm that the centered task vectors induced from weight averaging effectively reduce task interference than the original task vectors, enhancing the performance of model merging.
The proposed method, \ours{}, significantly outperforms previous merging approaches by a substantial margin across various tasks and modalities.
Moreover, our method can be easily plugged into any merging frameworks based on task arithmetic. 
We demonstrate this by incorporating \ours{} with test-time adaptation and show that it achieves performance comparable to individually fine-tuned models.

\clearpage
\paragraph{Acknowledgement}
This work was in part supported by the National Research Foundation of Korea (RS-2024-00351212, RS-2024-00436165), Institute of Information and communications Technology Planning and Evaluation (IITP) grant (RS-2024-00509279, RS-2022-II220926, RS-2022-II220959) funded by the Korean government (MSIT), and Artificial intelligence industrial convergence cluster development project funded by the Ministry of Science and ICT (MSIT, Korea) \& Gwangju Metropolitan City.

{
    \small
    \bibliographystyle{ieeenat_fullname}
    \bibliography{main}
}

\clearpage
\setcounter{page}{1}
\setcounter{section}{0}
\renewcommand*{\thesection}{\Alph{section}}

\maketitlesupplementary
\appendix

\section{Proof for Theorems}
\label{ref: orthogonality}
In here, we strcitly discuss the relationship between the Row Space Interference and the Task Interference for 1-layer neural network model.

Let \(T > 2\) be the number of tasks. For each task \(t \in [T]\), we have
\(
\mathcal{D}_t = \{(x_{t,i}, y_{t,i}) \}_{i=1}^{n},  x_{t,i},\, y_{t,i} \in \mathbb{R}^d,
\)
with the linear model
\(
y_{t,i} = (\theta_0 + \tau_t)x_{t,i}, \quad \theta_0,\, \tau_t \in \mathbb{R}^{d \times d}.
\)
Each task-specific matrix admits the SVD with non-zero singular values are bounded by some positive constant $[\alpha, s_{max}].$

We assume that inputs lie close to the subspace spanned by row space of $\tau_{t}$ \(V_t\):
\[
x_{t,i} = V_t\,a_{t,i} + \epsilon_{t,i}, \quad a_{t,i} \in \mathbb{R}^d,\; \|a_{t,i}\|_2 \leq c\cdot s_{\text{max}},\; \|\epsilon_{t,i}\|_2 \leq \eta
\] where $c$ is a positive constant and \(\eta\) is assumed to be a sufficiently small constant (i.e., \(\eta \ll c\cdot s_{\text{max}}\)).

\begin{theorem}
    \label{thm:interference_formal}
    Let a multi task model defined as $W_{\mathrm{MTL}} = \theta_0 + \sum_{t=1}^T \tau_t$ and
 Task Interference $L := \sum_{t=1}^T \sum_{i=1}^{n} \| W_{\text{MTL}} x_{t,i} - (\theta_0 + \tau _t) x_{t,i} \|_2^2$, then $L$ and Row space Interference I satisfies following bound: \begin{align}
    L \le n(k_3 I + T(T-1)k_4 \eta)^2
\end{align} where $k_3 = s_{\text{max}}^2 c \cdot \frac{r s_{\text{max}}^2}{\alpha^2}, \quad k_4 = s_{\text{max}}.$
\end{theorem}

\begin{proof}
By definition of $L$,
\begin{align*}
L = \sum_{t=1}^T \sum_{i=1}^{n} \left\| \sum_{s \neq t} \tau_s x_{t,i} \right\|_2^2.
\end{align*}

Since $x_{t,i} = V_t a_{t,i} + \epsilon_{t,i}$, we have:
\begin{align*}
\| \tau_s x_{t,i} \|_2 &\leq \| \tau_s V_t a_{t,i} \|_2 + \| \tau_s \epsilon_{t,i} \|_2 \\
&\leq s_{\text{max}} \| V_s^T V_t a_{t,i} \|_2 + s_{\text{max}} \| \epsilon_{t,i} \|_2 \\
&\leq s_{\text{max}} \| V_s^T V_t \|_2 \| a_{t,i} \|_2 + s_{\text{max}} \| \epsilon_{t,i} \|_2 \\
&\leq s_{\text{max}} (\| V_s^T V_t \|_2 c s_{\text{max}} + \eta).
\end{align*}

Thus, the norm of the sum over all $s \neq t$ is:
\begin{align*}
\left\| \sum_{s \neq t} \tau_s x_{t,i} \right\|_2 &\leq \sum_{s \neq t} \| \tau_s x_{t,i} \|_2 \\
&\leq \sum_{s \neq t} s_{\text{max}} (\| V_s^T V_t \|_2 c s_{\text{max}} + \eta).
\end{align*}

Squaring and summing over all $t$ and $i$, we get:
\begin{align*}
L &\leq \sum_{t=1}^T n \left( \sum_{s \neq t} s_{\text{max}} (\| V_s^T V_t \|_2 c s_{\text{max}} + \eta) \right)^2 \\
&\leq n \left( s_{\text{max}}^2 c \sum_{t=1}^T \sum_{s \neq t} \| V_s^T V_t \|_2 + T (T-1) s_{\text{max}} \eta \right)^2.
\end{align*}

Since $\tau_t$ can be composed $U_t \Sigma_t V_t^T$ by SVD with $\Sigma_t = \operatorname{diag}(s_{t,1}, \ldots, s_{t,r_t}, 0, \ldots, 0)$ where $r_t=rank(\tau_{t})$ and $s_{t,1} \ge s_{t,2} \ge ... \ge s_{t,rt} \ge 0$. Also, $\tilde{\Sigma}_t = \frac{\Sigma_t}{\| \tau_t \|_{\text{fro}}}$, where $\| \tau_t \|_{\text{fro}} = \sqrt{\sum_{k=1}^{r_t} s_{t,k}^2} \leq \sqrt{r_t} s_{\text{max}}$, by definition of $I$:
\begin{align*}
I &= \sum_{t=1}^T \sum_{s \neq t} \| V_t^T \tilde{\Sigma}_t \tilde{\Sigma}_s V_s \|_{\text{fro}} \\
&\geq \sum_{t=1}^T \sum_{s \neq t} \frac{\alpha^2}{\| \tau_t \|_{\text{fro}} \| \tau_s \|_{\text{fro}}} \| V_t^T V_s \|_2 \\
&\geq \sum_{t=1}^T \sum_{s \neq t} \frac{\alpha^2}{r_t s_{\text{max}}^2} \| V_t^T V_s \|_2.
\end{align*}

Since $r_t$ can be bounded by a counstant $r\le d$, we can bound:
\begin{align*}
I \geq \frac{\alpha^2}{r s_{\text{max}}^2} \sum_{t=1}^T \sum_{s \neq t} \| V_t^T V_s \|_2.
\end{align*}

Thus,
\begin{align*}
\sum_{t=1}^T \sum_{s \neq t} \| V_t^T V_s \|_2 \leq \frac{r s_{\text{max}}^2}{\alpha^2} I.
\end{align*}

Substituting back into the bound for $L$:
\begin{align*}
L &\leq n  \left( s_{\text{max}}^2 c \cdot \frac{r s_{\text{max}}^2}{\alpha^2} I + T(T-1) s_{\text{max}} \eta \right)^2 \\
&\le n \left( k_3 I + T(T-1) k_4 \eta \right)^2,
\end{align*}
where $k_3 = s_{\text{max}}^2 c \cdot \frac{r s_{\text{max}}^2}{\alpha^2}$, $k_4 = s_{\text{max}}$ and $r$ is max of $\{rank(\tau_{i})\}_{i=1}^{T}$.
\end{proof}



\newpage

\section{Implementation Details}
\paragraph{Searching the Scaling Coefficient and Rank}
In \ours{}, we performed a grid search for the scaling coefficient $\lambda \in (1.0, 0.2, 0.4, \ldots, 3.0)$ and selected $\lambda$ that yielded the best performance, following the task arithmetic settings~\cite{ilharco2023editing}. Also, The rank grid can be conducted within the range of $\{4\%, 8\%, 16\%, 32\%\}$, considering different tasks and modalities.
In CART++, which employs test-time adaptation to optimize the scaling coefficients, we adopted the settings from AdaMerging \cite{yang2024adamerging}.
Using the test dataset for each task, we iteratively optimized the scaling coefficients separately until the evaluation accuracy converged to an upper bound.
We provide the pseudocode for \ours{} and \ada{} in Algorithm~\ref{alg:full_algorithm}, where the test-time adaptation of \ada{} is described in Algorithm~\ref{alg:entropy_adaptation}.

\begin{algorithm}[t]
    \caption{\textbf{Model Merging with \ours{}}}
    \label{alg:full_algorithm}
    \begin{algorithmic}[1]
        \REQUIRE Fine-tuned parameters $\{\theta_t\}_{t=1}^T$, rank pruning ratio $\gamma$, scaling coefficient $\lambda$, test data $\mathcal{D}_{\text{test}}$ (optional)
        \ENSURE Merged model parameters $\theta_*$

        \FOR{each layer $l$}
        \STATE Compute the average of parameters:
        \begin{equation*}
            \theta_\text{avg}^{l} = \frac{1}{T} \sum_{t=1}^T \theta_t^{l}
        \end{equation*}

        \FOR{each task $t$}
        \STATE Compute the centered task vectors:
        \begin{equation*}
            \bar{\tau}_t^{l} = \theta_t^{l} - \theta_\text{avg}^{l}
        \end{equation*}
        \STATE Perform SVD: $\bar{\tau}_t^{l} = \sum_{i=1}^r \sigma_i \mathbf{u}_i \mathbf{v}_i^\top$
        \STATE Compute the pruning rank $k = \lceil r \times \gamma \rceil$
        \STATE Compute the low-rank approximation:
        \begin{equation*}
            \bar{\tau}_t^{l} \leftarrow \sum_{i=1}^k \sigma_i \mathbf{u}_i \mathbf{v}_i^\top
        \end{equation*}
        \ENDFOR
        \ENDFOR

        \IF{Test-Time Adaptation is enabled}
        \STATE \textbf{Call} \textbf{Algorithm \ref{alg:entropy_adaptation}} to optimize $\{\lambda_t^{l}
            \}$ using $\mathcal{D}_{\text{test}}$

        \COMMENTTRIANGLE{\ada{}}
        \ELSE
        \STATE Set $\lambda_t^{l} = \lambda$, \quad $\forall t, l$

        \COMMENTTRIANGLE{\ours{}}
        \ENDIF

        \FOR{each layer $l$}
        \STATE \textbf{Merge} models using optimized coefficients:
        \begin{equation*}
            \theta_*^{l} = \theta_\text{avg}^{l} + \sum_{t=1}^T \lambda_t^{l} \bar{\tau}_t^{l}
        \end{equation*}
        \ENDFOR
        \STATE \RETURN Merged model parameters $\theta_*$
    \end{algorithmic}
\end{algorithm}

\begin{algorithm}[t]
    \caption{\textbf{Entropy-Based Test-Time Adaptation}}
    \label{alg:entropy_adaptation}
    \begin{algorithmic}[1]
        \REQUIRE Test data $\mathcal{D}_{\text{test}}$, task vectors $\{\bar{\tau}_t^{l}\}$, average parameters $\theta_\text{avg}^{l}$, learning rate $\eta$, iterations $N$

        \STATE Initialize scaling coefficients $\{\lambda_t^{l}\}$
        \FOR{$n = 1$ \TO $N$}
        \FOR{each layer $l$}
        \STATE Compute merged model parameters:
        \begin{equation*}
            \theta_*^{l} = \theta_\text{avg}^{l} + \sum_{t=1}^T \lambda_t^{l} \bar{\tau}_t^{l}
        \end{equation*}
        \ENDFOR

        \STATE Compute entropy loss on test data:
        \begin{equation*}
            \mathcal{L}_{\text{entropy}} = -\frac{1}{|\mathcal{D}_{\text{test}}|} \sum_{\mathbf{x} \in \mathcal{D}_{\text{test}}} \sum_{c} p(c|\mathbf{x}; \theta_*) \log p(c|\mathbf{x}; \theta_*)
        \end{equation*}

        \FOR{each task $l$}
        \FOR{each task $t$}
        \STATE Compute gradient $\nabla_{\lambda_t^{l}} \mathcal{L}_{\text{entropy}}$
        \STATE Update scaling coefficients:
        \begin{equation*}
            \lambda_t^{l} \leftarrow \lambda_t^{l} - \eta \nabla_{\lambda_t^{l}} \mathcal{L}_{\text{entropy}}
        \end{equation*}
        \ENDFOR
        \ENDFOR
        \ENDFOR
        \STATE \RETURN Optimized scaling coefficients $\{\lambda_t^{l}\}$
    \end{algorithmic}
\end{algorithm}

\section{Additional Results}
\subsection{Further analysis on I}
\label{sec:anal I}
\begin{figure}[!t]
    \centering
    \includegraphics[width=0.9\linewidth]{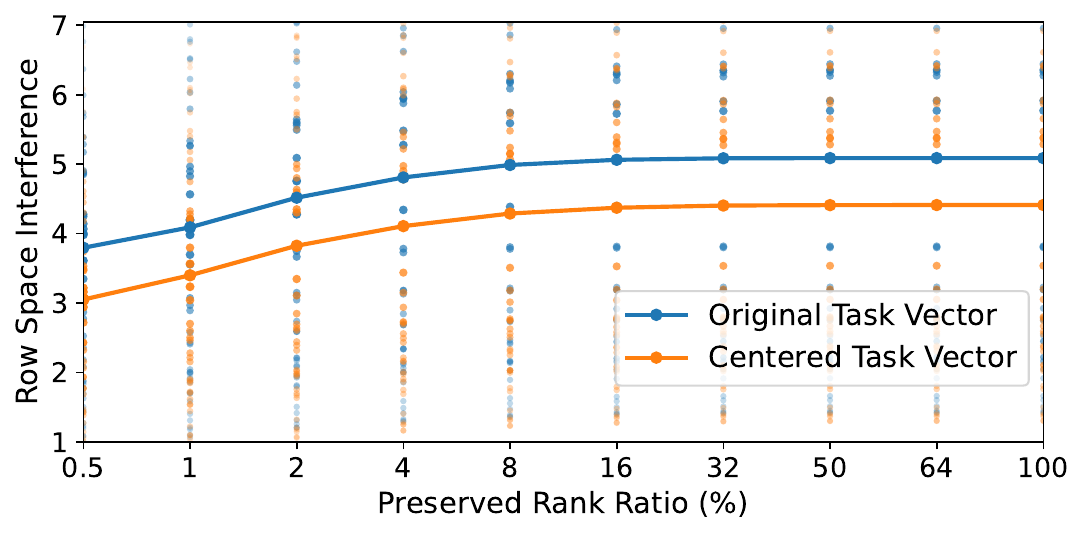}
    \caption{
        Scatter plot of Row Space interference $I(k)$ for each rank ratio across all layers, with blue and orange points representing Original Task Vector and Centered Task Vector, respectively. Point opacity and size reflect proximity to the median, while solid lines indicate the layer-wise median $I(k)$ for each method.
    }
    \label{fig:layer_wise_I}
\end{figure}

As depicted in Figure~\ref{fig:layer_wise_I}, our analysis demonstrates that centered task vector consistently shows low $I(k)$ across all layers than the original task vectors across all preserved rank ratios.

\subsection{Compression Performance}
\label{sec:compression}
\begin{table*}[ht]
      \small
      \centering
      \caption{Averaged absolute accuracy results on 8, 14 and 20 vision tasks model compresssion benchmark. For \oursindexing{}, we retain 8\% rank.}
      \begin{tabular}{ccccccc}
        \toprule
        \multirow{2}{*}{\textbf{Method}}         & \multicolumn{3}{c}{\vitb{}}                     & \multicolumn{3}{c}{\vitl{}}                    \\ 
        \cmidrule(lr){2-4} \cmidrule(lr){5-7}
                                                 & 8 tasks & 14 tasks          & 20 tasks           & 8 tasks & 14 tasks          & 20 tasks          \\ 
        \midrule
        \multicolumn{1}{c}{Pretrained}            & $48.0$ & $57.1$ & $56.0$ & $65.0$ & $68.4$ & $65.4$ \\
        \multicolumn{1}{c}{Individual}            & $90.5$ & $89.6$ & $90.5$ & $94.2$ & $93.3$ & $94.1$ \\
        \midrule
        \multicolumn{7}{c}{With Additional Per-task Parameters} \\
        \midrule
        \multicolumn{1}{c}{EMR-Merging}       & $88.8$ & $85.7$& $86.3$ & $93.5$& $91.7$& $92.0$\\
        \multicolumn{1}{c}{TALL Mask + Ties}          & $\mathbf{90.8}$ & $\mathbf{89.6}$& $\mathbf{90.2}$& $\mathbf{94.3}$& $92.4$& $93.1$\\
        \multicolumn{1}{c}{\textbf{\oursindexing{} (Ours)}}                 & $90.1$ & $89.2$& $90.1$& $93.9$& $\mathbf{93.1}$& $\mathbf{93.9}$\\
        \bottomrule
      \end{tabular}
      \label{tab:task_indexing}
    \end{table*}
\begin{table*}[ht]
    \small
    \centering
    \caption{Additional result of eight NLP tasks with model compression benchmark. For CART-Indexing, we retain 8\% rank.}
    \begin{tabular}{l | c c c c c c c c | c}
        \toprule
        Method           & CoLA   & SST2   & MRPC   & STSB   & QQP    & MNLI   & QNLI   & RTE    & \textbf{Avg.} \\
        \midrule
        Individual       & 0.6018 & 0.9404 & 0.8922 & 0.9063 & 0.9141 & 0.8720 & 0.9271 & 0.7906 & 0.8556        \\
        \midrule
        EMR-Merging & 0.3996 & 0.9335 & 0.8627 & 0.8277 & 0.8972 & 0.8545 & 0.8957 & 0.7437 & 0.8018        \\
        \textbf{\oursindexing{} (Ours)}  & \textbf{0.5832} & \textbf{0.9415} & \textbf{0.8897} & \textbf{0.9068} & \textbf{0.9001} & \textbf{0.8721} & \textbf{0.9249} & \textbf{0.7762} & \textbf{0.8493}        \\
        \bottomrule
    \end{tabular}
    \label{tab:roberta_index}
\end{table*}

If only minimal additional per-task parameters are allowed, \ours{} is even further applicable to compression‐based methods thanks to the low‐rank nature of our task vector. Specifically, for each task \(t\), we can perform inference as follows:
\[
    \tilde{\mathcal{A}(t)}_k = \theta_\text{avg}^l + \text{SVD}_k\bigl(\theta_t^l - \theta_\text{avg}^l\bigr), \quad \forall~ l \le L,
\]
which we refer to as \oursindexing{}. As demonstrated in Table~\ref{tab:task_indexing} and Table~\ref{tab:roberta_index}, our methodology exhibits performance comparable to the sophisticated compression techniques developed by \citet{huang2025emr} and \citet{wang2024localizing}.

For a network comprising \(L\) layers, each characterized by an \(M \times M\) weight matrix, and tasked with handling \(T\) tasks, the binary mask approach requires \(T \times L \times M^2\) bits for storage. In contrast, our method, which employs low-rank task vectors of rank \(k\), demands approximately \(32 \times T \times L \times 2Mk\) bits. Specifically, with \(k = 0.08\), our approach achieves a storage cost that is competitive when \(M\) is sufficiently large (e.g., \(M \approx 1024\)), and it becomes increasingly efficient for larger values of \(M\).

\subsection{Performance of Large Model}
\label{sec:large_model}
\begin{figure*}[!ht]
    \centering
    \includegraphics[width=1\linewidth]{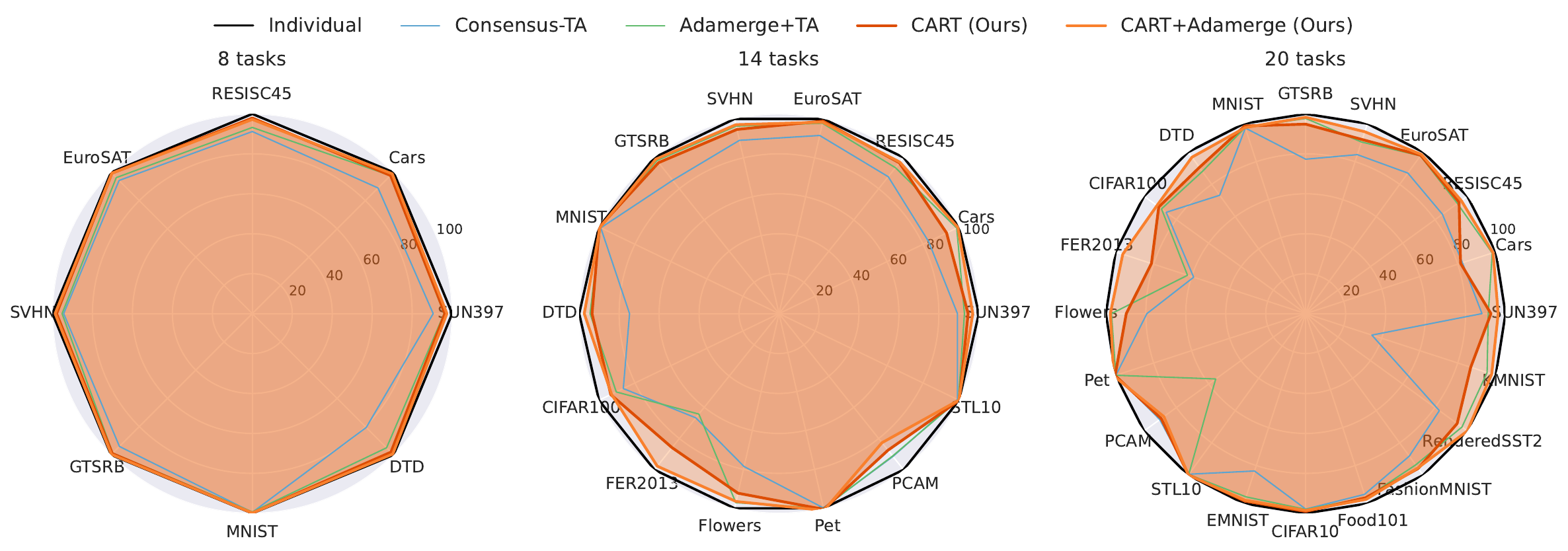}
    \caption{Radar plot of normalized accuracy results on 8, 14 and 20 vision classification tasks model merging benchmark for ViT-L-14.}
    \label{fig:radar_large}
\end{figure*}

\begin{figure*}[!t]
    \includegraphics[width=\textwidth]{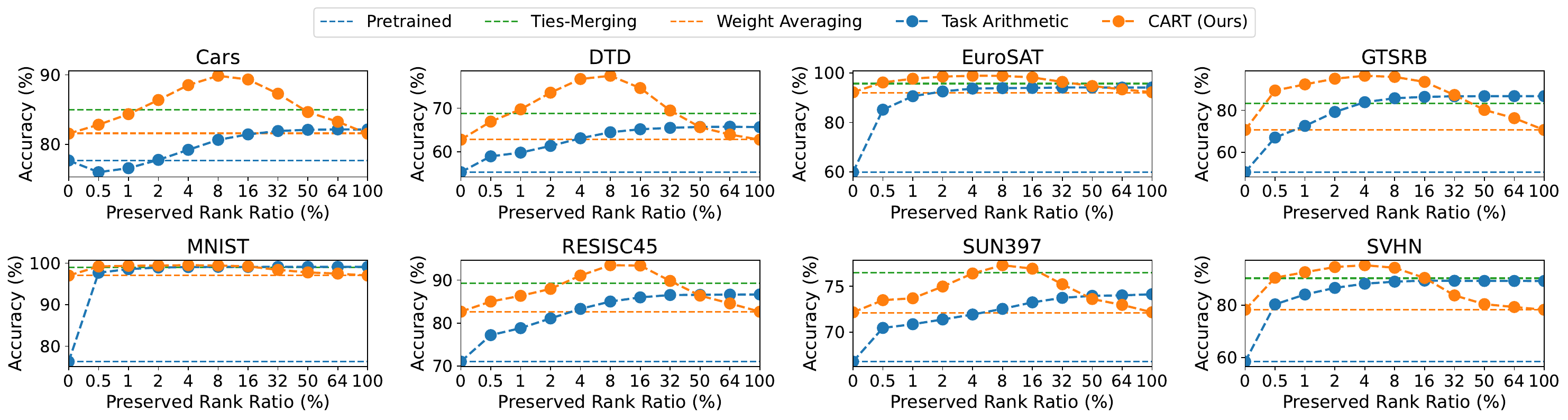}
        \caption{Performance plots for 8 vision task on \vitl{}.}
        \label{fig:perf_on_ranks_task_specific_large}
\end{figure*}

In this section, we provide Figure~\ref{fig:perf_on_ranks_task_specific_large} that demonstrate the optimal rank for merging lies around $8\%$ of the full rank for \vitl{} as \vitb{} shows. 
Also, we provide the detailed performance of \ours{} for \vitl{} in vision classification benchmarks as Figure~\ref{fig:radar_large}.

\section{Sampling methods for Task Number scalability Experiments}
\label{sec:task_sampling}
To obtain Figure~\ref{fig:performance_diff_number_tasks}, we adopted a sampling strategy, selecting a subset of combinations from the 20 vision tasks for each \( N \).
This approach was necessitated by the computational infeasibility of evaluating all possible combinations of \( N \) tasks out of 20, as the total number of combinations, denoted \( \binom{20}{N} \), becomes prohibitively large (e.g., \( \binom{20}{8} = 125,970 \)). To circumvent this challenge, we relied on the Central Limit Theorem, which asserts that, given a sufficiently large sample size, the distribution of the sample mean approximates a normal distribution, irrespective of the underlying population distribution. Accordingly, for a specified confidence level of $95\%$, a population standard deviation \( \sigma \), and a desired margin of error \( \epsilon \) in estimating the mean performance, a sample size of at least
\begin{equation}
    m = \left\lceil \left( \frac{1.96 \cdot \sigma}{\epsilon} \right)^2 \right\rceil 
\end{equation}
combinations is sufficient to ensure an accurate estimate of the true mean. Here, \( \sigma \) represents the standard deviation of the performance metric across all possible combinations, and the value 1.96 corresponds to the critical value \( z_{0.025} \) from the standard normal distribution for a 95\% confidence level. In practice, since \( \sigma \) is typically unknown, we employed the Popoviciu inequality  to estimate the $\sigma$, which provides an upper bound on the variance of a bounded random variable. For a random variable \( X \) constrained to the interval \( [a, b] \), the Popoviciu inequality asserts that the variance \( \text{Var}(X) \) satisfies:
\begin{equation*}
    \text{Var}(X) \leq \frac{(b - a)^2}{4}
\end{equation*}
This implies that the standard deviation \( \sigma \) is bounded by:
\begin{equation}
    \sigma \leq \frac{b - a}{2}
\end{equation}
In our experimental setup, we applied this inequality to establish a upper bound for \( \sigma \) for each value of \( N \). Specifically, we defined the lower bound \( a \) as the average of the bottom \( N \) zero-shot performance values across the 20 tasks, reflecting the lowest-performing baseline configurations. Similarly, we set the upper bound \( b \) as the average of the top \( N \) fine-tuned model performance values, capturing the highest-performing baseline configurations. By determining \( a \) and \( b \) in this manner, we tailored the upper bound of \( \sigma \) to each \( N \), ensuring that the sample size \( m \) is sufficiently large so that the sample mean reliably estimates the population mean.

To determine hyperparameters for each merge method, we used the default values or recommended hyperparameters presented in each merge method. Specifically, for Task Arithmetic, as defined in Equation~\ref{eqn:task_arithmetic}, we performed a grid search for the merge coefficient $\lambda$ over the range $[0.1, 0.2, 0.3]$ to identify the optimal value for each task combination. For Ties-Merge, we performed a grid search for the parameter $k$ over $[5, 10, 20]$, where $k$ represents the percentage of top magnitude parameters to retain in the task vectors, setting the remaining parameters to 0, and fixed the merge coefficient $\lambda = 1.0$. For our proposed method, we performed a grid search for the low rank ratio over $[0.08, 0.12, 0.16]$, and fixed the merge coefficient $\lambda = 1.0$.

\section{Further Analysis on Loss Landscapes}

\begin{figure*}[!t]
    \centering
    \includegraphics[width=1\linewidth]{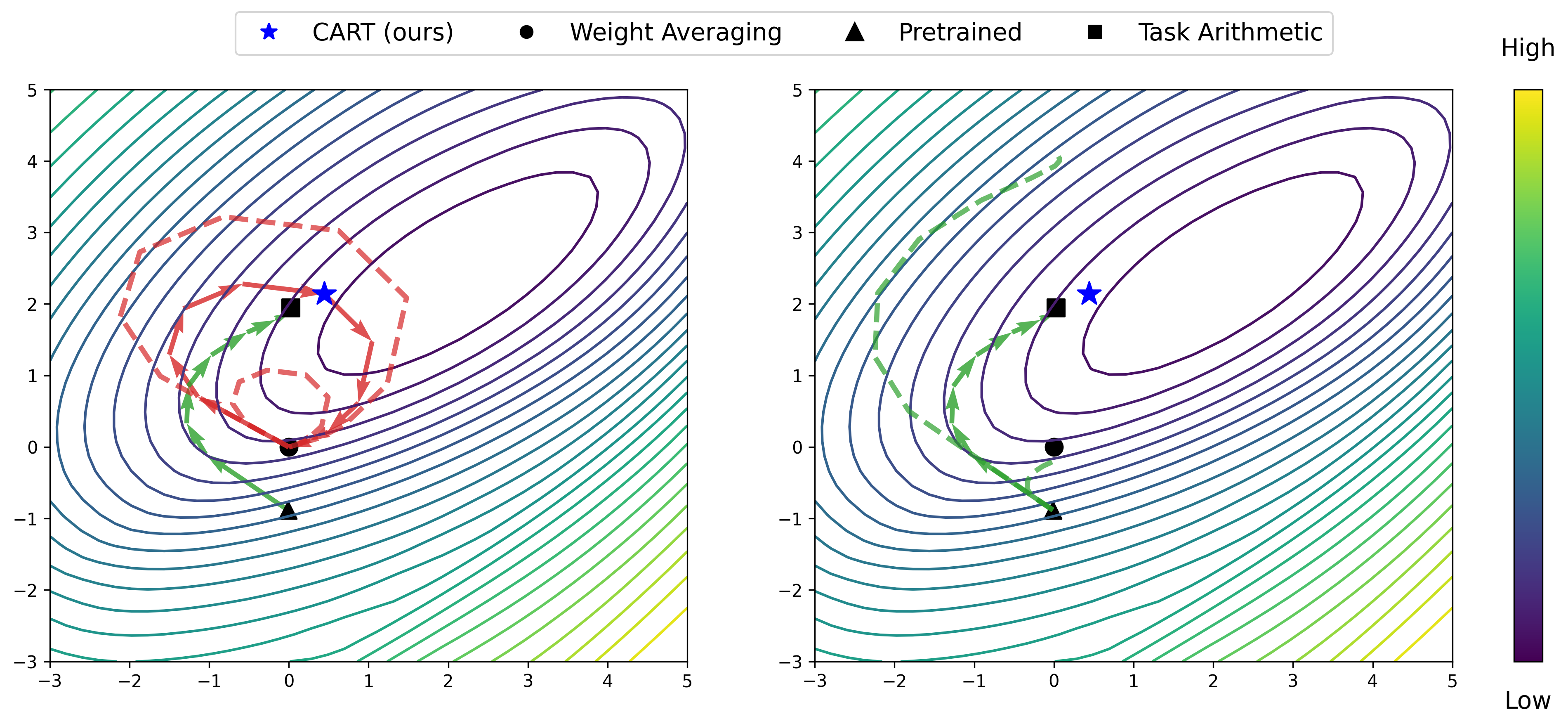} 
    \caption{
        We present a visualization of the multi-task loss landscape (i.e., $\sum_{t} \mathcal{L}_{t}(\theta)$) for \ours{} and task arithmetic using the ViT-B-32 model with varying values of $k$. For both model merging methodologies, we plot the parameter trajectories by incrementally increasing $k$ from 0 to the full rank. Specifically, in our approach, the trajectory begins and returns to the weight average $\theta_{\text{avg}}$, represented by a red line. In contrast, task arithmetic's trajectory progresses from the pretrained model to the merged weights, depicted by a green line. Additionally, the black circle indicates the weight average $\theta_{\text{avg}}$, the blue star marks \ours{} with 8\% of the full rank, the black triangle represents the pretrained model, and the black square denotes the merged model obtained through task arithmetic. Trajectories corresponding to different $\lambda$ values are illustrated as dashed lines.
    }
    \label{fig:mt_loss}
\end{figure*}

\paragraph{Multi-Task Loss Landscape}
To directly compare the behaviors of \ours{} and task arithmetic in the parameter space, we visualize the loss landscapes of them with respect to the multi-task loss (\emph{e.g.,} $\mathcal{L} = \sum_{t=1}^T \mathcal{L}_t$).
As observed in the left panel of Figure~\ref{fig:mt_loss}, \ours{} can closely approach the common basin induced by the multi-task loss, with its trajectory rotating from $\theta_{\text{avg}}$ as the rank increases.
In contrast, for task arithmetic, we observed that even as the rank increases (up to full rank), it converges to a point farther from the common basin than \ours{}, which is consistent with Figure~\ref{fig:perf_on_ranks}.
In the right panel of Figure~\ref{fig:mt_loss}, we examine the trajectory of task arithmetic with varying $\lambda$. We observe that when $\lambda = 0.125$, the trajectory moves toward the weight average, and as $\lambda$ increases (\emph{e.g.,} to 1), there is a tendency to move away from the common basin.

\begin{figure*}[!ht]
    \centering
    \includegraphics[width=1\linewidth]{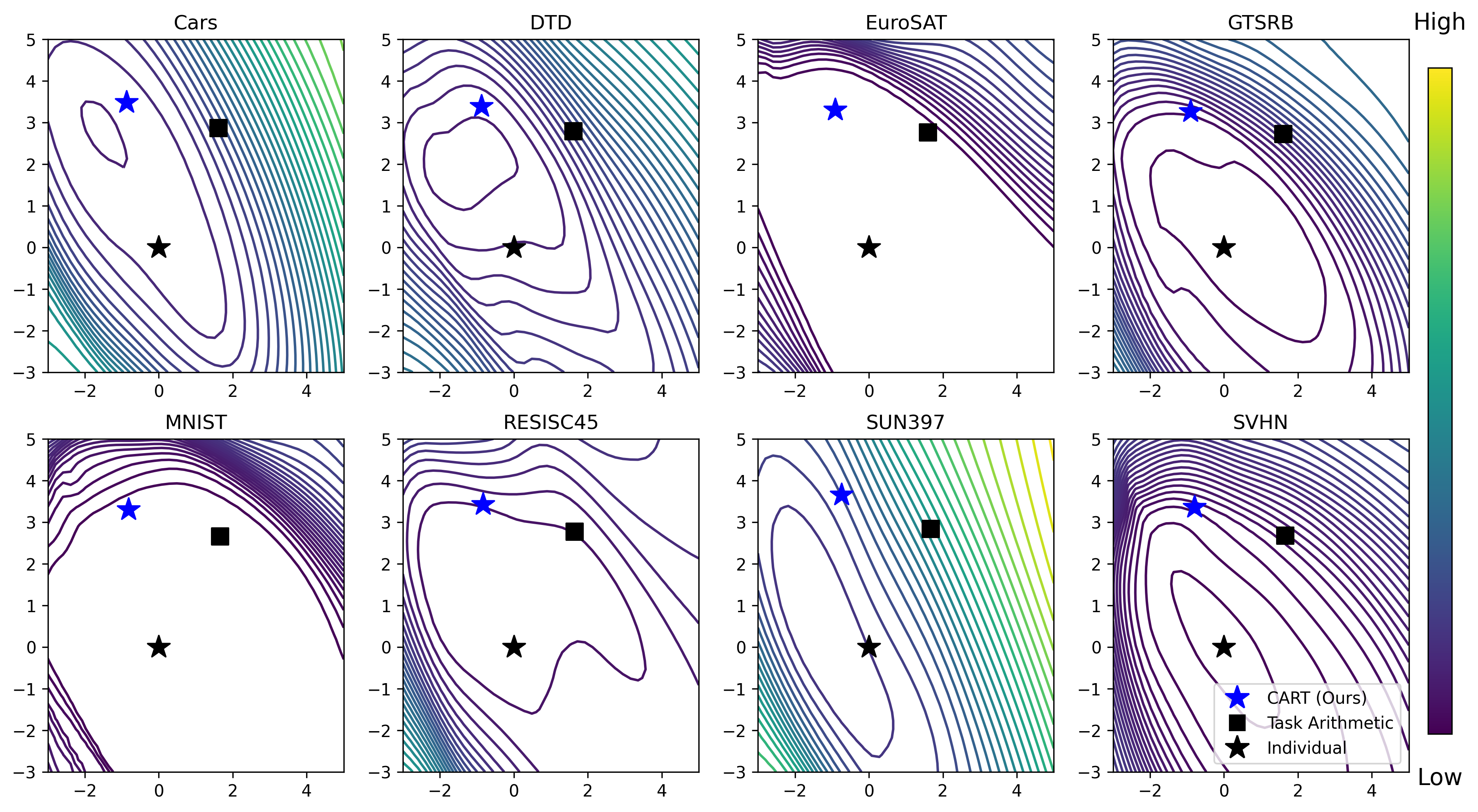}
    \caption{Visualizations of the loss landscape surrounding the individually fine-tuned models for each task using the ViT-B-32 model. The black star represents the individual parameters $\theta_t$, the blue star denotes \ours{} with 8\% of the full rank, and the black square indicates the merged model obtained through task arithmetic.
    }
    \label{fig:loss_landscape_with_single_task}
\end{figure*}

\paragraph{Task-wise Loss Landscapes}
In Figure~\ref{fig:loss_landscape_with_single_task}, we investigate the loss landscape around the fine-tuned weights $\theta_{t}$ for each task. We confirm that the $\theta_{t}$ are located within the basin for most tasks. Additionally, by examining the positions of \ours{} and task arithmetic, we observe that \ours{} is closer to the loss basin of the single tasks, which is consistent with the results shown in Figure~\ref{fig:perf_on_ranks_task_specific_large}.

\end{document}